# Logical Semantics and Commonsense Knowledge: Where Did we Go Wrong, and How to Go Forward, Again

**Walid S. Saba**

Astound.ai
111 Independence Drive, Menlo Park, CA 94025
walid@astound.ai

**Abstract**

We argue that logical semantics might have faltered due to its failure in distinguishing between two fundamentally very different types of concepts: ontological concepts, that should be types in a strongly-typed ontology, and logical concepts, that are predicates corresponding to properties of and relations between objects of various ontological types. We will then show that accounting for these differences amounts to the integration of lexical and compositional semantics in one coherent framework, and to an embedding in our logical semantics of a strongly-typed ontology that reflects our commonsense view of the world and the way we talk about it in ordinary language. We will show that in such a framework a number of challenges in natural language semantics can be adequately and systematically treated.

## Introduction

In the concluding remarks of *Ontological Promiscuity* Hobbs (1985) made what we believe to be a very insightful observation: given that semantics is an attempt at specifying the relation between language and the world, if "one can assume a theory of the world that is isomorphic to the way we talk about it ... then semantics becomes nearly trivial". But how exactly can we rectify our logical formalisms so that semantics, an endeavor that has occupied the most penetrating minds for over two centuries, can become (nearly) trivial, and what exactly does it mean 'to assume a theory of the world' in our semantics?

In this paper we hope to provide answers for both questions. First, we believe that a commonsense theory of the world can (and should) be embedded in our semantic formalisms resulting in a logical semantics grounded in commonsense metaphysics. Furthermore, we believe that the first step to accomplishing this vision starts by rectifying what we think was a crucial oversight in logical semantics, namely the failure to distinguish between two fundamentally different types of concepts: ontologica1l concepts, that are **types** in a strongly-typed ontology; and logical concepts, that are **predicates** corresponding to properties of and relations between objects of various ontological types. This differentiation, as we hope to show, results in a natural linking between ordinary spoken language and commonsense knowledge about the world we talk about[1]. In particular, by embedding ontological types in our predicates type unification and other type operations can then be used to 'uncover' missing information – information that is never explicitly stated in everyday discourse, but is often implicitly assumed as shared background knowledge.

In the next section we briefly discuss the phenomenon of the 'missing text', which is behind most challenges in the semantics of natural language. We will then suggest how accounting for the difference between logical and ontological concepts coupled with type unification over predicates embedded with ontological types can help us 'uncover' this missing text. Subsequently, it will be shown how in such a framework a number of challenges in the semantics of natural language can be adequately and uniformly treated. Finally, we will briefly discuss the nature of the strongly-typed ontological structure that will be assumed throughout the paper and argue that this structure cannot be 'invented' but should be 'discovered', briefly making some suggestions on how this might be accomplished.

## The Phenomenon of the 'Missing Text'

Perhaps for computational effectiveness, as Givon (1984) once suggested, in using ordinary spoken language to express our thoughts we tend to do so by using the least possible effort; by, for one thing, uttering the least number of words that are needed to convey a particular thought. Thus, for example, we make statements such as:

(1) a. *Simon is a rock*.
 b. *The ham sandwich wants a beer*.

---

[1] Throughout this paper reference to a 'commonsense theory of the world' is a reference to the commonsense knowledge required in the language understanding process (as performed by a 5-year old), and not to specialized domain knowledge that one would need in general problem solving.

c. *Sheba is articulate.*
   d. *Jon bought a brick house.*
   e. *Carlos likes to play bridge.*
   f. *Jon visited a house on every street.*
   g. *Jon has Das Kapital but he never read it.*

In our opinion, speakers (hearers) of ordinary language utter (understand) these sentences to convey (mean) the following, respectively:

(2) a. *Simon is* [**as solid as**] *a rock*.
   b. *The* [**person eating the**] *ham sandwich wants a beer.*
   c. *Sheba is* [**an**] *articulate* [**person**].
   d. *Jon bought a brick* [**-made**] *house.*
   e. *Carlos likes to play* [**the game**] *bridge.*
   f. *Jon visited a* [**different**] *house on every street.*
   g. *Jon has* [**the book**] *Das Kapital but he never read it*[**'s content**].

Clearly, any viable semantic formalism must somehow 'uncover' and account for this [**missing text**], as such sentences are quite common and are not at all exotic, or farfetched. In this regard, we are in total agreement with Levesque (2011), who states that in order to comprehend such sentences, "you need to have background knowledge that is not expressed in the words of the sentence to be able to sort out what is going on … And it is precisely bringing this background knowledge to bear that we informally call *thinking*" (emphasis in original)[2]. However, and although these sentences seem to have a common denominator, it is somewhat surprising that in looking at the literature one finds that these phenomena have been studied quite independently and in many cases with incompatible proposals that are individually tailored to a specific phenomenon, such as metaphor (2a); metonymy (2b); textual entailment (2c); nominal compounds (2d); lexical ambiguity (2e); quantifier scope ambiguity (2f); reference resolution and salient meanings (2g); and copredication (2h), to name a few. In recent years, however, there has been quite a bit of work to deal with some instances of this phenomenon by incorporating type systems in logical semantics. For example, in a series of papers (see e.g. Lou, 2011; Lou, 2011; Lou 2012) Lou introduced a type system based on Martin-Lof's type theory (Martin-Lof, 1984) where common nouns are considered to be types, and where it is shown how the machinery of *type coercion* can in such a system handle lexical disambiguation and accommodate for copredication; where the latter refers to situations where we have a 'structured object' (or a dot-type) object that can be predicated in different ways in the same context (see Pustojosky, 1995). An example of this is given in (3), where there is a reference to the PHYSICAL-OBJECT sense of 'book' (when being bought) and to the INFORMATIONAL-CONTENT aspect of book (when being read):

(3) *John bought and read the latest book on deep learning*

While we are sympathetic to the general approach of Lou, we believe that copredication and lexical ambiguity are part of a single and much simpler phenomenon (which we will shortly get into), and thus we believe that type coercion introduces complex machinery unnecessarily, not to mention that type shifting/coercing will not always produce the desired results. The same observation can be made about the work of Asher and Pustejovsky (2012) where complex machinery that permits type shifting is also used to access different aspects (senses) of a structured object (a dot-type) using the lexical constraints available in the context. The problem we have with this approach is that the notion of the dot-type does not seem to be cognitively plausible since language allows us to refer to many aspects of a given object that cannot all be a priori defined as part of the lexical semantics. Moreover, and more specific to the work of Lou, we argue that there is in fact a technical problem in assuming that the entire class of common nouns should constitute the types in the system. Consider for example the sentence in (4).

(4) *John is an excellent teacher*

Clearly 'excellent' in (4) does not describe *John*, but his teaching activity (formally, *John is excellent* does not follow from *John is an excellent teacher*), and thus it is the hidden teaching `activity` that is the ontological type here, and not the common noun, 'teacher'[3].

Starting with (Asher, 2008; Asher, 2011) and more recently in (Asher, 2015), Asher has also developed over a few years a type system that uses type coercion to account for lexical disambiguation as well as to handle situations involving copredication. While the same reservations we have regarding the approach taken by Lou (2012) more or less apply to the earlier work of Asher, the more recent Asher (2015) however correctly highlights the technical problems in simply performing type shifting (or type coercion), and in particular in examples such as those in (5).

(5) *Julie enjoyed a book. It was a mystery*

If a straightforward type shifting is performed on the first sentence, so that the type constraints imposed by 'enjoy' (which expects an eventuality) are satisfied, then the subsequent sentence cannot be correctly interpreted as we would have lost, so to speak, the physical object sense of book that is the obvious referent of 'it'. Asher concludes, and correctly so, in our opinion, that it is not type shifting

---

[2] Note that the only plausible explanation to the fact that we all tend to leave out the same information is that the background knowledge needed to comprehend such sentences must be *shared*.

[3] The same argument applies to situations where the common noun refers to an event, a state, a property, etc. (more on this below)

of book that must occur, but that some process in predicate composition must occur. That process, which Asher (2015) calls 'transformation', is essentially a functor that 'picks-up' the desired object that can semantically link to the verb's argument. We are in general agreement with the spirit of this approach (as nothing else will work, for one thing), but we have two reservations. For one thing, this 'transformation' operation is not very clear, especially in how it picks-up different kinds of objects, for example, an EVENTUALITY (that is a 'reading') in (6) and an INFORMATIONAL CONTENT object in (7):

(6) *Julie enjoyed the book.*
    $\Rightarrow$ *Julie enjoyed **reading** the book.*
(7) *Julie criticized the book.*
    $\Rightarrow$ *Julie criticized **the content of** the book.*

Moreover, and if types are meant to be a set of general categories (e.g. PHYSICAL-OBJECT, INFORMATIONAL-CONTENT, ANIMATE, etc.), it is not clear how it can be determined in (8) that it was Barcelona's *residents* who voted for independence (8a), that it was Barcelona's *team* that lost to Real Madrid (8b), and that it was Barcelona's *governing body* that announced a curfew (8c).

(8) *Barcelona was calm after it* $\begin{cases} \text{a. voted for independence} \\ \text{b. lost to Real Madrid} \\ \text{c. announced a curfew} \end{cases}$

The reason the highly general (and somewhat ad-hoc) type system is potentially problematic is that the missing terms in (8) (VOTERS, SPORTS-TEAM, GOVERNING-BODY) are, at a high level, of a similar type, namely some 'group of people' and thus for our pragmatics to work it must operate at a much more granular level.

In summary, we want to point out that recent efforts to incorporate type-theory in compositional semantics, thus integrating lexical semantics (and commonsense metaphysics) with compositional semantics is a welcoming trend, and the pioneering work of Pustejovsky (2012), Lou (2011) and Asher (2015) are efforts in the right direction. However, and as will be suggested below, these specific proposals to treat a few phenomenon in the semantics of natural language seem to be ad-hoc and do not address the more general problem of how in the process of language *understanding* we seem to be able to uncover all the implicitly assumed but missing information. In the rest of this paper we will offer an alternative approach that we believe will answer this question. In particular, we will argue that most of the challenges in the semantics of natural language (lexical disambiguation, copredication, metonymy, etc.) are due to the phenomena of the 'missing text' - information that is never explicitly stated but is often implicitly assumed as shared background knowledge. As such, we will show that uncovering such information requires rectifying a crucial oversight in logical semantics, namely the uniform use of predication to represent two fundamentally different types of concepts: ontological concepts (that represent types in a strongly-typed ontology) and logical concepts (that are properties of and relations between objects of various ontological types).

## Ontological vs. Logical Concepts

Before we start with our proposal, let us briefly suggest where we believe the standard logical treatment of natural language semantics might have faltered. Consider the sentences in (9) and their translations to standard first-order predicate logic in (10).

(9) a. *Julie is an articulate person*
    b. *Julie is articulate*
(10) a. PERSON(*julie*) ∧ ARTICULATE(*julie*)
     b. ARTICULATE(*julie*)

We argue that, from a cognitive and a commonsense point of view, the translation of (9a) to that of (10a) is problematic. First, note that (9a) and (9b) have, more or less, the same cognitive content, and thus (10a) and (10b) must accordingly have more or less the same truth value conditions. But that effectively means that the truth value of ARTICULATE(*julie*) is only meaningful if PERSON(*julie*) is considered to be true a priori. Stated in other words this means that ARTICULATE(*julie*) is meaningful only if *Julie* is assumed a priori to be an object of **type** person. Since $a \wedge b \equiv b \wedge a$ this sequential order is not captured in (10a), however. The problem in our opinion is that PERSON is not a predicate, but a type that is presupposed by the logical concept (the property) ARTICULATE. Thus, we argue that the proper translation of both sentences in (9) is the one given in (11).

(11) (∃*julie* :: **person**)(ARTICULATE(*julie*))

That is, the proper translation is one that states that 'there is an object named *julie*, an object of type **person**, such that the property ARTICULATE is true of *julie*. In the next section we expand on this idea in more details.

## Types vs. Predicates

In *Types and Ontology* Fred Sommers (1963) suggested years ago that there is a strongly typed ontology (that he termed 'the language tree') that seems to be implicit in all that we say in ordinary spoken language, where two objects *x* and *y* are considered to be of the same type iff the set of monadic predicates that are significantly (that is, truly or falsely but not absurdly) predicable of *x* is equivalent to the set of predicates that are significantly predicable of *y*. Thus, for example, while the noun phrases in (12) make reference to two distinct sets of objects, for an ontologist interested in the relationship between ontology and language, the noun phrases in (12) are ultimately referring to one type only, namely **cat**:

(12)  a. *an old cat*
      b. *a black cat*

In other words, whether we make a reference to an *old cat* or to a *black cat*, in both instances we are ultimately speaking of objects that are of the same type; and this, according to Sommers, is a reflection of the fact that the set of monadic predicates in our natural language that are significantly predicable of 'old cats' is exactly the same set that is significantly predicable of 'black cats' (or, whatever can sensibly be said of black cats can also be sensibly said of old cats, and vice versa). In this sense, a concept such as OLD is a predicate that happens to be predicable of a concept such as **cat**, which corresponds to a type in a strongly-typed ontology. As such, we take the proper logical representation for the noun phrase in (13) to be that in (13b), and not the one in (13a).

(13)  ⟦*an adorable cat*⟧
      a. $\Rightarrow \lambda P[(\exists x)(\text{CAT}(x) \land \text{ADORABLE}(x) \land P(x))]$
      b. $\Rightarrow \lambda P[(\exists x :: \mathtt{cat})(\text{ADORABLE}(x) \land P(x))]$

That is, *an adorable cat* refers to some object of type **cat**, a cat that, presumably, is adorable[4]. Note also that abstract objects, such as events, states, properties, are also types in the ontology that can be predicated, as shown in (14).

(14)  ⟦*an imminent event*⟧
      $\Rightarrow \lambda P[(\exists x :: \mathtt{event})(\text{IMMINENT}(x) \land P(x))]$

      ⟦*an idle state*⟧ $\Rightarrow \lambda P[(\exists x :: \mathtt{state})(\text{IDLE}(x) \land P(x))]$

      ⟦*a desirable property*⟧
      $\Rightarrow \lambda P[(\exists x :: \mathtt{property})(\text{DESIRABLE}(x) \land P(x))]$

In our representation, therefore, we assume a Platonic universe that includes everything we talk about in our language, and where concepts belong to two quite distinct categories: (*i*) ontological concepts, such as **animal**, **substance**, **entity**, **artifact**, **book**, **event**, **state**, etc., which are types in a subsumption hierarchy, and where the fact that an object of type **human** is (ultimately) an object of type **entity** is expressed as $\mathtt{human} \leqslant \mathtt{entity}$; and (*ii*) logical concepts, such as FORMER, OLD, IMMINENT, BEAUTIFUL, etc., which are the properties (that can be said) of and the relations (that can hold) between ontological concepts. The following are examples that illustrate the difference between logical and ontological concepts:

(15)  $R_1$: OLD($x$ :: **entity**)
      $R_2$: HEAVY($x$ :: **physical**)
      $R_3$: HUNGRY($x$ :: **living**)
      $R_4$: ARTICULATE($x$ :: **human**)
      $R_5$: MAKE($x$ :: **human**, $y$ :: **artifact**)
      $R_6$: MANUFACTURE($x$ :: **human**, $y$ :: **instrument**)
      $R_7$: RIDE($x$ :: **human**, $y$ :: **vehicle**)
      $R_8$: DRIVE(X :: **human**, y :: **car**)

The predicates in (15) are supposed to reflect the fact that in ordinary spoken language we can say OLD of any **entity**; that we say HEAVY of objects that are of type **physical**; that HUNGRY is said of objects that are of type **living**; that ARTICULATE is said of objects that must be of type **human**; that make is a relation that can hold between a **human** and an **artifact**; that MANUFACTURE is a relation that can hold between a **human** and an **instrument**, etc. Note that the type assignments in (15) implicitly define a type hierarchy as that shown in figure 1 below. Consequently, and although not explicitly stated in (15), in ordinary spoken language one can always attribute the property HEAVY to an object of type **car** since $\mathtt{car} \leqslant \mathtt{vehicle} \leqslant \mathtt{physical}$[5]. In addition to logical and ontological concepts, there are also proper nouns, which are the names of objects that could be of any type. A proper noun, such as *Sheba*, for example, is interpreted as follows ⟦*Sheba*⟧ $\Rightarrow \lambda P[(\exists^1 Sheba :: \mathtt{thing})(P(x))]$.

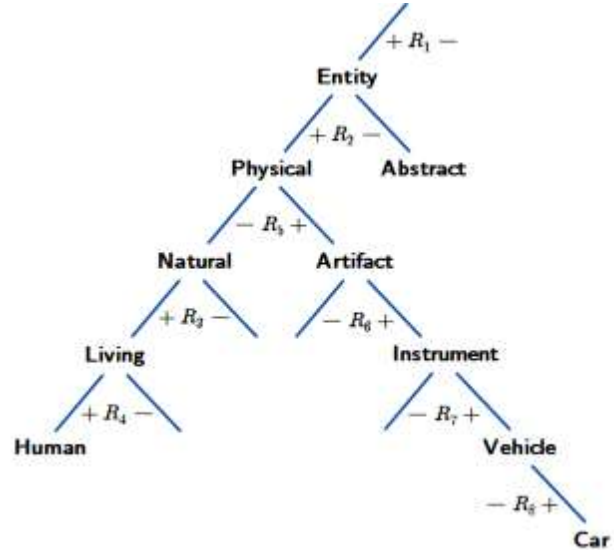

**Figure 1.** The type hierarchy implied by (15)

---

[4] As Hacking (2001) suggests one can think of the type as **cat** the kind of object that answers a question such as 'What-is-it?' The distinction between types and predicates is thus related to the analytic/synthetic distinction, where the truth of a type judgment such as (*sheba*::**cat**) is a synthetic judgment the truth of which is determined by virtue of what we know about the world; while the truth of the judgment WILD(*sheba*::**cat**) is determined by virtue of the meaning of 'wild'. As such, and while currently popular data-driven approaches to natural language are in our opinion totally misguided, type judgments like ($x$::**cat**) probably do belong to the quantitative level in that the truth of the type judgment ($x$::**cat**) could be determined by pattern recognition systems, including type judgments of abstract objects (does some event 'look like' a dancing **event**, etc.)

[5] It should be noted here that the expressions in (15) are assumed to refer to a specific sense of each predicate. In general, however, the type assignment is a set of possible types where a single type is eventually left after lexical disambiguation. This will be discussed in more detail below.

A point worth mentioning at this early juncture is that besides the embedding of 'commonsense' constraints in our predicates, what implicitly gets defined by applying Sommers' predicability test, as given by (15), is the implicit determination of 'saliency'. For example, and while it makes sense to speak of **human** objects that MAKE, RIDE and DRIVE objects of type **car**, DRIVE is a more salient relation between a **human** and a **car** since a **human** rides a car as a **vehicle**, and makes a car as an **artifact**, but s/he drives a car explicitly as a **car** (see figure 2).

**Type Unification**

Let us now start our 'compositional' semantics. Consider the interpretation of *Sheba is a thief* where we assume THIEF is a property that is ordinarily said of objects that must be of type **human**, that is THIEF($x$ :: **human**):

(16)   ⟦*Sheba is a thief*⟧ ⇒
        ($\exists^1$*Sheba* :: **thing**)(THIEF(*Sheba* :: **human**))

Thus, *Sheba is a thief* is interpreted as: there is a unique object named *Sheba* and initially assumed to be of type **thing**, such that the property THIEF is true of *Sheba*[6].

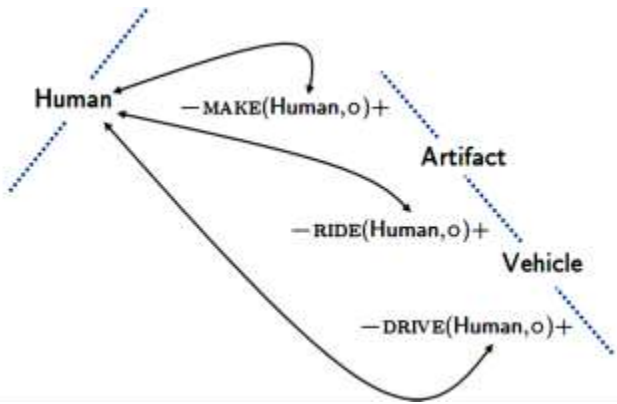

**Figure 2.** Salient relations implied by types and their properties.

Note now that in (16) *Sheba* is associated with more than one type in a single scope. In these situations a type unification must occur, where a type unification (**s** ● **t**) between types **s** and **t** and where $Q \in \{\exists, \forall\}$ is defined, for now, as follows:

(17)
$Q(x :: (\mathbf{s} \bullet \mathbf{t}))(P(x))$
$\equiv \begin{cases} Q(x :: \mathbf{s})(P(x)), & if(\mathbf{s} \leqslant \mathbf{t}) \\ Q(x :: \mathbf{t})(P(x)), & if(\mathbf{t} \leqslant \mathbf{s}) \\ Q(x :: \mathbf{s})Q(x :: \mathbf{t})(P(x) \wedge \mathbf{R}(x)), & if(\exists \mathbf{R})(\mathbf{R} = msr(\mathbf{s}, \mathbf{t})) \\ \bot, & otherwise \end{cases}$

---

[6] For simplicity, we are ignoring for now some intermediate steps in the translation, especially as it relates to the copula 'is' (more on this below).

where *msr*(**s**,**t**) stands for the most salient relation between objects of type **s** and objects of type **t**. That is, in situations where there is no subsumption relation between **s** and **t** the type unification results in keeping the variables of both types and in introducing some salient relation between the two types (we will discuss these situations below). Going to back to (16), the type unification in this case is actually quite simple, since (**human** ⩽ **thing**):

(18)   ⟦*Sheba is a thief*⟧ ⇒
        ($\exists^1$*Sheba* :: **thing**)(THIEF(*Sheba* :: **human**))
        ($\exists^1$*Sheba* :: (**thing** ● **human**))(THIEF(*Sheba*))
        ($\exists^1$*Sheba* :: **human**)(THIEF(*Sheba*))

In the final analysis, therefore, *Sheba is a thief* is interpreted as follows: there is a unique object named *Sheba*, an object that eventually came out to be of type **human**, such that THIEF is true of *Sheba*. Note the clear distinction between ontological concepts (e.g., **human**), which Cocchiarella (2001) calls first-intension concepts, and logical (or second-intension) concepts, such as THIEF($x$ :: **human**). In accordance with Quine's famous dictum ("to be is to be the value of a variable"), what (18) says is that what ontologically exist are objects of type **human**, and not thieves, and THIEF is an accidental (as well as temporal, etc.) property that we came to use to talk of certain objects of type **human**. Furthermore, it is assumed that a logical concept such as THIEF is defined by a logical expression such as ($\forall x ::$ **human**)(THIEF($x$) ≡ $\phi$), where the exact nature of $\phi$ might very well be susceptible to temporal, cultural, and other contextual factors depending on what, at a certain point in time, a certain community considers a THIEF to be.

It should also be noted that a first-intension such as ($x ::$ **human**) as well as a second-intension such as ARTICULATE($x ::$ **human**) are both 'judgments' where the former is a type judgment and the latter is a value judgment. As such, for a judgment such as ARTICULATE($x ::$ **human**) to be made, the type judgment ($x ::$ **human**) must first be made. The way we see it, therefore, is that type judgments are the first level in the overall semantic structure, as shown in figure 3.

**More on Type Unification**

Consider the following interpretation of *Sara owns a black cat*, where we assume that BLACK can be said of objects of type **physical**, and that the OWN relationship holds between objects of type **human** and objects of type **entity**:

(19)   ⟦*Sara owns a black cat*⟧ ⇒
        ($\exists^1$*Sheba* ::**thing**)($\exists c$::**cat**)
           (BLACK($c$ :: **physical**)
              ∧ OWN(*Sara* :: **human**, c :: **entity**))

Thus *Sara owns a black cat* is initially interpreted as: there is a unique **thing** named *Sara* and an object *c* of type **cat**

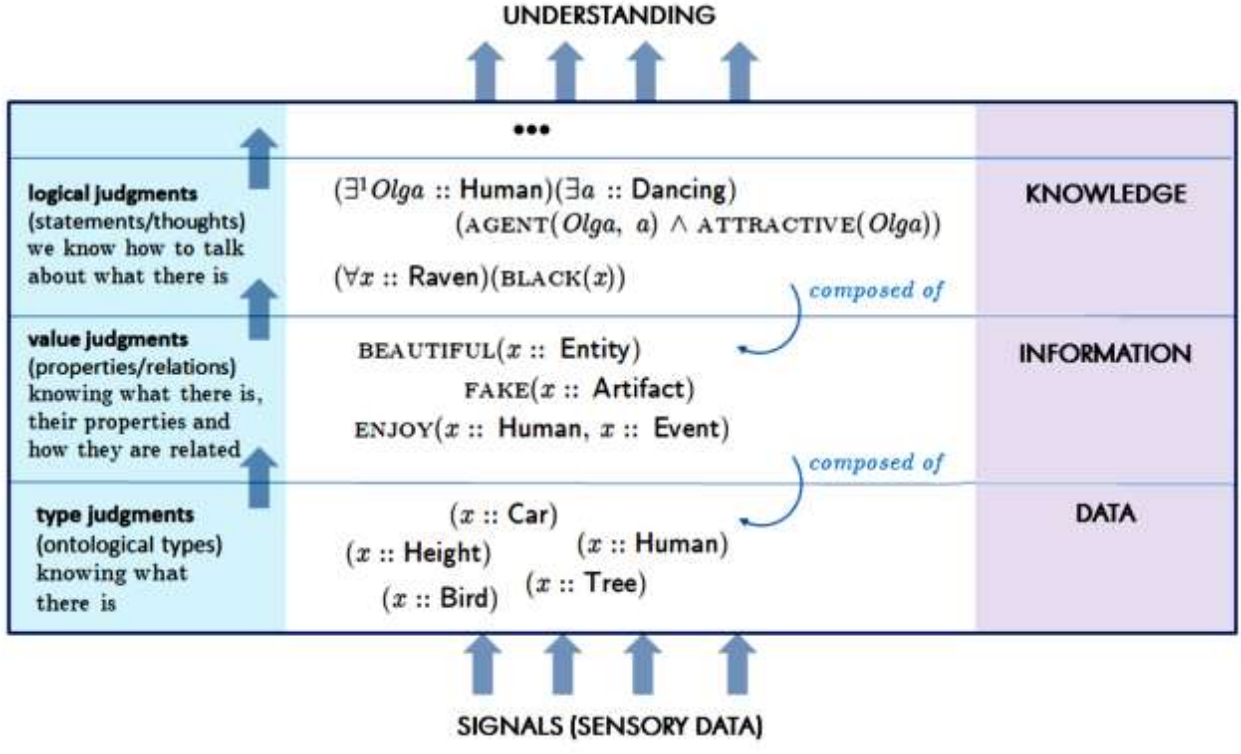

**Figure 3.** Levels of understanding

such that *c* is BLACK (and thus in this context it must be of type **physical**), and *Sara* owns *c*, where in this context *Sara* must be object of type **human** and *c* an object of type **entity**. Depending on the context they are mentioned in, therefore, *Sara* and *c* are assigned different types. The type unifications that must occur in this situation are the following (where '→' means 'unifies to'):

(*Sara* :: (**thing** • **human**)) → (*Sara* :: **human**)
(*c* :: ((**physical** • **entity**) • **cat**))
→ (*c* :: (**physical** • **cat**))
→ (*c* :: **cat**)

The final interpretation of *Sara owns a black cat* is finally given by the following:

(20)  ⟦*Sara owns a black cat*⟧ ⇒
(∃$^1$*Sheba*::**human**)(∃*c*::**cat**)(BLACK(*c*) ∧ OWN(*Sara*,*c*))

That is, there is unique object named *Sara*, which is of type **human**, and some object *c* of type **cat**, where *c* is BLACK and *Sara* owns *c*.

**Type Unification and Abstract Objects**

As discussed above, logical concepts such as TEACHER, THIEF, etc. are assumed to be defined by some logical expression. A plausible definition for a logical concept such as DANCER could for example be given by (21).

(21)  (∀*x*::**human**)
(DANCER(X) ≡ (∃*a*::**dancing**)(AGENT(*a*,*x*))

That is, any *x* (that must be of type **human**) is a DANCER iff *x* is the agent of some **dancing** (a subtype **activity**). Let us now consider the interpretation of *Olga is a beautiful dancer* where we assume BEAUTIFUL(*a*::**entity**) – i.e., 'beautiful' is a property that can be said of any **entity**:

(22)  ⟦*Olga is a beautiful dancer*⟧ ⇒
(∃$^1$*Olga* :: **thing**)(∃$^1$*a* :: **dancing**)
(AGENT(*a* :: **activity**, *Olga* :: **human**) ∧
(BEAUTIFUL(*Olga*::**entity**) ∨ BEAUTIFUL(*a*::**entity**)))

Thus, *Olga is a beautiful dancer* is initially translated as follows: there is a unique **thing** named *Olga*, and some **dancing** *a*, where *Olga* is the agent of *a*, which must be an **activity** (and as the agent, *Olga* must be of type **human**), and where either *Olga* is beautiful or her **dancing** (or, of course, both). Note now that *Olga* and *a* are assigned three types in the same scope, triggering the following type unifications:

(*Olga* :: ((**thing** • **entity**) • **human**))
→ (*Olga* :: (**entity** • **human**)) → (*Olga* :: **human**)

(*a* :: ((**dancing** • **activity**) • **entity**))
→ (*a* :: (**dancing** • **entity**)) → (*a* :: **dancing**)

Concerning the disjunction term in (14), representing the ambiguity in nominal modification, we now have the following BEAUTIFUL(*Olga*::**human**) ∨ BEAUTIFUL(*a*::**dancing**)). Since both terms in the disjunction are acceptable, the final translation, admitting the ambiguity in the nominal modification, is the one given in (23).

(23) ⟦*Olga is a beautiful dancer*⟧ ⇒
 (∃¹*Olga* :: **human**)(∃a :: **dancing**)
  (AGENT(*a*,*Olga*) ∧ (BEAUTIFUL(*Olga*) ∨ BEAUTIFUL(a)))

Unlike the situation in (23), however, the relevant type unifications should remove the ambiguities in (24) and (25),

(24) *Olga is an experienced dancer*
(25) *Olga is a recreational dancer*

where it is clear that *experienced* is describing *Olga* in the former and *recreational* is describing *Olga*'s dancing in the latter. The term we need to reconsider here is the term involving the disjunction representing the ambiguity in nominal modification. The type unifications in the case of (24):

EXPERIENCED(*Olga* :: (**human** • **human**))
  ∨ EXPERIENCED(*a* :: (**dancing** • **human**))
→ EXPERIENCED(*Olga* :: **human**) ∨ EXPERIENCED(*a* :: ⊥)
→ EXPERIENCED(*Olga* :: **human**) ∨ ⊥
→ EXPERIENCED(*Olga* :: **human**)

The type unification admitting an 'experienced dancing' fails here, leaving 'experienced' to unambiguously modify *Olga*. In (25), however, we have the following:

RECREATIONAL(*Olga* :: (**human** • **dancing**))
  ∨ RECREATIONAL(*a* :: (**dancing** • **dancing**))
→ RECREATIONAL(*Olga* :: ⊥) ∨ RECREATIONAL(*a*::**dancing**)
→ ⊥ ∨ RECREATIONAL (*a* :: **dancing**)
→ RECREATIONAL(*a* :: **dancing**)

Note that in this case 'recreational Olga' was eliminated leaving 'recreational' to unambiguously modify *Olga*'s **dancing**. A valid question here is this: why was the type unification (**dancing** • **human**) in (24) and (25) considered a failure (resulting in ⊥), although the definition of type unification given in (17) suggests that in the absence of a subsumption relation an attempt is first made is to pick-up the 'most salient relation' (*msr*) between the two ontological types. The answer is that looking for an *msr* occurs when all else *locally* fails, while this is not the case in (24) and (25) where the context provided a successful type unification and thus looking elsewhere to 'make sense' of what is being said was not needed!

### An Innate Ontological Structure?

A longstanding subject of debate in linguistics is the phenomenon of adjective-ordering restrictions (AORs), which is concerned with the apparent preferred adjective orderings we tend to have when multiple adjectives are used in a sequence. For example, it is generally agreed that the ordering of adjectives in (26a) is preferred to that in (26b).

(26) a. *Jon bought a beautiful red car*
    b. *Jon bought a red beautiful car*

What makes this an interesting phenomenon from a cognitive science point of view is the apparent universality and cross-linguistic nature of AORs, as well as the fact that children seem to effortlessly make these preferences without ever being instructed on what the most 'natural' ordering is. Various studies have suggested that these ordering preferences might be a function of some syntactic and semantic classes of adjectives (e.g., Cinque 1994; Larson, 1998; and Vendler, 1968), nouniness – how close is the adjective to a noun, or temporariness (how much does the adjective encode a temporary property) (e.g., Teodorescu, 2006). Despite all these studies, the debate as to what might explain these ordering preferences, and whether the ordering preferences might reflect a much deeper cognitive phenomenon, have not yet been settled. What we like to suggest here is that adjective-ordering restrictions, are related to type casting in a strongly-typed ontological structure. For example, the adjective orderings in (27a) and (27b) require the following type unifications:

(27) a. BEAUTIFUL(RED(*x* :: **physical**) :: **entity**)
    b. RED(BEAUTIFUL(*x* :: **entity**) :: **physical**)

Note that the type unification in (27a) requires a type casting from **physical** to **entity**, while the one in (27b) requires a type casting in the reverse order. The type casting in (27b) should be blocked, however, since (as it is well known in type theory of programming languages) one can always perform type casting upwards, but not downwards[7].

What is interesting to contemplate here is this: if the adjective-ordering restriction phenomenon (which seems to be universally and cross-linguistically valid) turns out to be explainable by type casting and unification in a strongly-type ontological structure, it would suggest that the ontological structure we envision might have some innate underpinnings, as recent psycholinguistic research seems to suggest that this might be the case (Scontras et. al., 2017).

---
[7] Technically, we can always generalize by ignoring specific details; casting down (assuming unknown details), is however undecidable.

### Interaction between Type Casting and Unification

Recall the interpretation of *Olga is a beautiful dancer* in (23), where the final interpretation admitted the ambiguity in nominal modification where all type unifications succeeded, allowing 'beautiful' to remain ambiguous in that it could be modifying *Olga* or her dancing. Consider now the interpretation of *Olga is a beautiful tall dancer*, where we assume TALL($x$ :: **physical**):

(28) ⟦*Olga is a beautiful tall dancer*⟧ $\Rightarrow$
    ($\exists^1 Olga$ :: **thing**)($\exists a$ :: **dancing**)
      (AGENT($a$ :: **activity**, *Olga* :: **human**) $\wedge$
        (BEAUTIFUL(TALL(*Olga* :: **physical**) :: **entity**)
          $\vee$ BEAUTIFUL(tall($a$ :: **physical**) :: **entity**)))

The type relevant unifications here are the following,

(*Olga* :: ((**thing** • **physical**) • **human**))
$\rightarrow$ (*Olga* :: (**physical** • **human**))
$\rightarrow$ (*Olga* :: **human**)

($a$ :: ((**dancing** • **activity**) :: **physical**))
$\rightarrow$ ($a$ :: (**dancing** • **physical**))
$\rightarrow$ ($a$ :: $\bot$)
$\rightarrow$ $\bot$

resulting in (29):

(29) ⟦*Olga is a beautiful tall dancer*⟧ $\Rightarrow$
    ($\exists^1 Olga$ :: **human**)($\exists a$ :: **dancing**)(AGENT($a$,*Olga*)
      $\wedge$ (BEAUTIFUL(TALL(*Olga* :: **physical**))
        $\vee$ BEAUTIFUL(TALL($a$ :: $\bot$) :: **entity**))) $\Rightarrow$

    ($\exists^1 Olga$ :: **human**)($\exists a$ :: **dancing**)
      (AGENT($a$,*Olga*) $\wedge$ (BEAUTIFUL(TALL(*Olga*)) $\vee$ $\bot$)) $\Rightarrow$

    ($\exists^1 Olga$ :: **human**)($\exists a$ :: **dancing**)
      (AGENT($a$,*Olga*) $\wedge$ BEAUTIFUL(TALL(*Olga*)))

Unlike the situation in (23), where 'beautiful' could be describing *Olga* or her dancing, the situation in (29) is quite different due to the adjective 'tall' that cast the type of 'beautiful' to a **physical**, an thus to describe *Olga*, and not her dancing. We leave it to the reader to work out why (and how) *Olga is a beautiful tall dancer* sounds fine, while *Olga is a tall beautiful dancer* sounds awkward (hint: in the latter, the type castings and type unifications reduce the entire disjunction to $\bot$).

## Where Logical Semantics Can (Should?) Go

In this section we show how the embedding of ontological concepts in our predicates can help us tackle some well-known challenges in the semantics of natural language.

### Lexical Disambiguation

Thus far we have been assuming single type assignments to predicates variables, e.g. BLACK($x$ :: **physical**), as it was implicitly assumed that the predicate in question has been disambiguated – that is, that a specific meaning of the predicate has been selected. We will continue to do so where the context is clear, although we will show hear how lexical disambiguation itself is conducted in our system, and that requires that we initially consider, for some terms, a set of type assignments.

Let us consider the interpretation of the sentence in (30), where we assume 'party' has two meanings: the 'social event' and the 'political group'.

(30) ⟦*Jon cancelled the party*⟧ $\Rightarrow$
    ($\exists^1 Jon$ :: **thing**)($\exists a$ :: **activity**)
      ($\exists^1 p$::{**politicalGroup**, **socialEvent**})
        (CANCELLATION(a) $\wedge$ SUBJECT($a$, *Jon* :: **human**))
          $\wedge$ OBJECT($a$, $p$ :: **event**))

Note that, initially, the type associated with 'the party' $p$ is a set of all possible types (again, for simplicity we assumed that 'cancelled' has been disambiguated where it was determined that it is some activity the object of which is of type **event**). The type unifications for *Jon* are straightforward: (*Jon* :: (**human** • **thing**)) $\rightarrow$ (*Jon* :: **human**). The type unifications that must occur for $p$, however, are now a set of $n$ pairs of type unifications, where $n$ is the number of possible meanings of $p$:

($p$::{(**event** • politicalGroup), (**event** • **socialEvent**)})
$\rightarrow$ ($p$ :: {$\bot$, **socialEvent**})
$\rightarrow$ (p :: {**socialEvent**})

Thus, the initial set of types is reduced to a singleton and the 'party' that *Jon* seems to have cancelled is a 'social event'. Note that if 'cancelled' in (30) was replaced by 'assisted', then the correct meaning of 'party' will also be selected, namely the meaning of the political group. On the other hand, if 'cancelled' where to be replaced by 'promoted' then we would have a genuinely ambiguous statement, since one can 'promote' a political group, as well as a social event, as illustrated by (31).

(31) ⟦*Jon promoted the party*⟧ $\Rightarrow$
    ($\exists^1 Jon$ :: **thing**)($\exists a$ :: **activity**)
      ($\exists^1 p$ :: {**politicalGroup**, **socialEvent**})
        (PROMOTION($a$) $\wedge$ SUBJECT($a$, *Jon* :: **human**))
          $\wedge$ OBJECT($a$, $p$ :: **entity**))

Assuming that any **entity** can be promoted, and in the absence of any additional information both assumed meanings of 'party' remain to be equally plausible.

## Metonymy, Copredication and Salient Meanings

Consider the sentence in (32), where two senses of 'book' are assumed to be used in the same context, the informational **content** sense of book (when being read) and the **physical** object sense (when being burned):

(32) *Jon read the book and then he burned it.*

In Asher and Pustejovsky (2005) it is argued that 'book' must have what is called a dot type, which is a structured object that in a sense carries the 'informational content' sense (which is referenced when it is being read) as well as the 'physical object' sense (which is referenced when it is being burned). Elaborate machinery is then introduced to 'pick out' the right sense in the right context, and all in a well-typed compositional logic. But this approach presupposes that one can enumerate, a priori, all possible uses of the word 'book' in ordinary language. What we suggest, instead, is that 'copredication' is not different from metonymy, as in both cases we are trying to 'pick-up' for some concept *a* some other unmentioned concept *b* along with the most salient relation between them. To illustrate this point further, let us first consider the interpretation of *Jon bought and studied Das Kapital*:

(33) ⟦*Jon bought and studied Das Kapital*⟧ ⇒
 ($\exists^1 Jon$ :: **thing**)($\exists^1 DasKapital$ :: **book**)
  ($\exists a_1$ :: **activity**)($\exists a_2$ :: **activity**)
   (STUDYING($a_1$) ∧ SUBJECT($a_1$, $Jon$ :: **human**)
    ∧ OBJECT($a_1$, *DasKapital* :: **infContent**)
    ∧ BUYING($a_2$) ∧ SUBJECT($a_2$, $Jon$ :: **human**)
    ∧ OBJECT($a_2$, *DasKapital* :: **physical**))

That is, *Jon* bought a book that he also studied, where the object of the buying activity must be an object of type **physical**, and the object of a studying is some **infContent**. The type unifications of *Jon* and the **physical** book *DasKapital* are straightforward:

($Jon$ :: (**human** • **thing**))
→ ($Jon$ :: **human**)

(*DasKapital* :: (**physical** • **book**))
→ (*DasKapital* :: **book**)

However, the object of the studying **activity**, which is a **book**, must be an **infContent** object when being read. In the absence of 'local' options, for these two types to unify the most salient relation between **book** and **infContent** is picked-up, introducing in the process a new variable of type **infoc** that is related to a book: HASCONTENT(*DasKapital*, $x$ :: **infoc**). The final translation is thus:

(34) ⟦*Jon bought and studied Das Kapital*⟧ ⇒
 ($\exists^1 Jon$ :: **human**)($\exists^1 DasKapital$ :: **book**)
 ($\exists a_1$ :: **activity**)($\exists a_2$ :: **activity**))($\exists x$ :: **infoc**) ∧
  (STUDYING($a_1$) ∧ SUBJECT($a_1$, $Jon$) ∧ OBJECT($a_1$, $x$) ∧
   BUYING($a_2$) ∧ SUBJECT($a_2$, $Jon$) ∧ OBJECT($a_2$, *DasKapital*) ∧
   HASCONTENT(*DasKapital*, $x$))

That is, *Jon bought and studied Das Kapital* describes a situation where there is a unique object named Jon, an object of type **human**, and some **book** titled *DasKapital* (that Jon bought), and where *DasKapital* has the **infContent** $x$ (that *Jon* studied).

It is important to note at this stage that hidden (and implicitly assumed) information is either obtained by straight forward type unification or, when all attempts fail, by picking up some salient property or relation between the objects in the discourse. On the other hand, unwanted meanings (as in lexical disambiguation, or removing some ambiguities related to nominal modification) are obtained when certain type unifications fail.

Another important point that we like to make here is that copredication, the name given for the phenomenon exemplified by (34), is not much different from what is known by metonymy, in that type unification is the process by what which an indirect reference or some salient relation are discovered. Consider for example the following:

(34) ⟦*The omelet wants a beer*⟧ ⇒
 ($\exists^1 oml$ :: **omelet**)($\exists b$ :: **beer**)($\exists a$ :: **activity**)
  (WANTING($a$) ∧ SUBJECT($a$, $oml$ :: **human**)
   ∧ OBJECT($a$, $b$ :: **thing**))

In this case, resolving the situation of 'what is wanted' is quite simple: the object of wanting is an object of type **thing**, and more specifically, a **beer**, which works very well: ($b$ :: (**beer** • **thing**)) → ($b$ :: **beer**). However, it is the subject of the wanting that seems to be the problem: 'want' expects a **human** subject but we found an object of type **omelet**. Clearly, these two types must somehow be reconciled. Since no subsumption relation exists between these types, an attempt at finding some salient relation between them is made. As it turns out, there is a salient relationship between a **human** and **food** (a supertype of **omelet**), namely the eat relation, that will necessarily introduce an (implicit) object of type **human**. Thus,

(35) ⟦*The omelet wants a beer*⟧ ⇒
 ($\exists^1 oml$ :: **omelet**)($\exists b$ :: **beer**)($\exists a_1$ :: **activity**)
  ($\exists a_2$ :: **activity**)($\exists x$ :: **human**)
   (EATING($a_2$) ∧ SUBJECT($a_2$, $x$) ∧ OBJECT($a_2$, $oml$)
    WANTING($a_1$) ∧ SUBJECT($a_1$, $x$) ∧ OBJECT($a_1$, $b$))

That is, there is some EATING and some WANTING that is going on (by some 'discovered' $x$), and the object of the EATING is an **omelet**, and the subject of the EATING, who is a human, wants a **beer**.

## The Ontology of Natural Language

Throughout this paper we have assumed the existence of some ontological structure, an ontological structure the types of which are assumed to be embedded in predicates (the properties of and the relations between objects of various types). However, a valid question that one might ask is the following: how does one arrive at this ontological structure that implicitly underlies all that we say in everyday discourse? One plausible answer is the (seemingly circular) suggestion that the semantic analysis of natural language should itself be used to uncover this structure. In this regard we strongly agree with Dummett (1991) who states:

> We must not try to resolve the metaphysical questions first, and then construct a meaning-theory in light of the answers. We should investigate how our language actually functions, and how we can construct a workable systematic description of how it functions; the answers to those questions will then determine the answers to the metaphysical ones.

What this suggests, and correctly so, in our opinion, is that in our effort to understand the complex and intimate relationship between ordinary language and everyday (commonsense) knowledge, one could, as Bateman (1995) has also suggested, ''use language as a tool for uncovering the semiotic ontology of commonsense'' since language is the only theory we have of everyday knowledge. To alleviate this seeming circularity (in assuming this ontological structure in our semantic analysis; while at the same time suggesting that semantic analysis of language should itself be used to uncover this ontological structure), we suggest performing semantic analysis from the ground up, assuming a minimal (almost a trivial and basic) ontology, building up the ontology as we go guided by the results of the semantic analysis. The advantages of this approach are: (i) the ontology thus constructed as a result of this process would not be invented, as is the case in most approaches to ontology (e.g., Guarino, 1995, Lenat & Guha, 1990, and Sowa, 1995), but would instead be discovered from what is in fact implicitly assumed in our use of language in everyday discourse; (ii) the semantics of several natural language phenomena should as a result become trivial, since the semantic analysis was itself the source of the underlying knowledge structures (in a sense, one could say that the semantics would have been done before we even started!)

Another promising technique that can be used to 'boot-up' this ontological structure is performing some corpus analysis to initially obtain sets of monadic predicates that seem to be (sensibly) used in the predication of some very general types (e.g., **artifact**, **event**, **physical**, **state**, etc.) A subset relationship analysis can then be used to discover the hidden hierarchical structure. A similar approach has been suggested in Saba (2010).

## Concluding Remarks

Most of the challenges in the semantics of natural language seem to be related to the phenomenon of the 'missing text' – that is, text is almost never explicitly stated but is implicitly assumed as shared background knowledge. In this paper we suggested how such commonsense knowledge might be 'uncovered' in the process of language *understanding*. In particular, we suggested that the answer lies in rectifying what we believe was an oversight in logical semantics, namely not distinguishing between two fundamentally different types of concepts: (*i*) ontological concepts that correspond to types in a strongly-typed ontology; and (*ii*) logical concepts, that correspond to the properties of and the relations between objects of various ontological types. We showed how in such a framework a number of challenges in the semantics of natural language can be adequately and systematically treated.

Several other linguistic phenomena intensionality (e.g., intensional verbs, intensional adjectives, etc.), or reasoning with abstract objects compound nominals, quantifier scope ambiguity, etc., were not dealt with in this paper as that would require some minor extension to our formalism and would thus extend the paper considerably. These and a more detailed presentation of some of the subjects covered in this paper are forthcoming. Much is left to be done, refined, and restated, of course, especially as it relates to the 'discovery' of that ontological structure that seems to be implicit in everything we say in our everyday discourse.